\documentclass[11pt]{article}

\usepackage[final]{acl}

\usepackage{times}
\usepackage{latexsym}
\usepackage{xspace}
\usepackage[table]{xcolor}
\usepackage{amsmath}
\usepackage{url}

\usepackage[T1]{fontenc}
\usepackage[utf8]{inputenc}

\usepackage{microtype}

\usepackage{inconsolata}

\usepackage{graphicx}
\usepackage{booktabs}
\usepackage{multirow}
\usepackage{soul}
\sethlcolor{yellow}

\def \method{\textsc{RPT}\xspace}
\newcommand{\gepaconf}{GEPA-C\xspace}

\title{Reflective Prompt Tuning through Language Model Function-Calling}

\author{
 \textbf{Farima Fatahi Bayat},
 \textbf{Moin Aminnaseri},
 \textbf{Pouya Pezeshkpour},
 \textbf{Estevam Hruschka}
\\
 Megagon Labs
\\
    \{farima, moin, pouya, estevam\}@megagon.ai
\\
}

\definecolor{cadmiumgreen}{rgb}{0.0, 0.42, 0.24}
\definecolor{cardinal}{rgb}{0.77, 0.12, 0.23}
\definecolor{cadmiumred}{rgb}{0.89, 0.0, 0.13}

\usepackage[most]{tcolorbox}
\newtcolorbox[list inside=prompt,auto counter,number within=section]{prompt}[1][]{
    fontupper=\ttfamily\footnotesize,
    boxsep=5pt,
    left=0pt,
    right=0pt,
    top=0pt,
    bottom=0pt,
    boxrule=1pt,
    breakable,
    #1,
}

\newtcolorbox{example}[2][]{%
  enhanced, breakable,
  colback=white,           % white background
  colframe=black!10,
  boxrule=0.4pt,
  coltitle=black,
  fonttitle=\bfseries,
  title={#2},
  #1
}

\begin{document}
\maketitle
\begin{abstract}
Large language models (LLMs) have become increasingly capable of following instructions and complex reasoning, making prompting a flexible interface for adapting models without parameter updates.
Yet prompt design remains labor-intensive and highly sensitive to formatting, phrasing, and instruction order, motivating automated prompt optimization methods that reduce manual effort while preserving inference-time flexibility.
However, existing methods often search over prompt candidates or use fixed critique-refine pipelines driven by individual examples or small batches, limiting their ability to capture systematic error patterns and make targeted edits grounded in failure history.
We propose Reflective Prompt Tuning (\method), a framework that uses LLM function calling to simulate the iterative workflow of human prompt engineers.
An LLM optimizer calls a diagnostic function that evaluates the target model over an entire optimization set, summarizes recurring failure modes, and returns a structured diagnostic report.
The optimizer uses this report, together with an accumulated memory of prior reports, to revise the prompt for the next iteration.
\method further supports confidence-aware optimization by using calibration signals in diagnostic feedback and final prompt selection.
Across three reasoning tasks, \method improves over initial prompts by up to 12.9 points, remains competitive with state of the art, and improves confidence calibration.
Our analyses show that \method is especially effective on multi-hop and mathematical reasoning, producing targeted prompt revisions that align with diagnosed failure patterns and lead to gains in task performance and calibration.\footnote{We release
our code at: \url{https://github.com/megagonlabs/RPT}.}
\end{abstract}

\section{Introduction}
Large language models (LLMs) have become increasingly adept at following instructions and performing complex reasoning, making contextual prompting the dominant mechanism for adapting model behavior to downstream tasks~\cite{10.1162/coli_a_00523, weicot2022, 10.5555/3600270.3601883}. Prompts let users specify objectives, constraints, and output formats without modifying model parameters, enabling rapid adaptation across applications~\cite{sahoo2025systematicsurveypromptengineering, schulhoff2025promptreportsystematicsurvey}.

Despite this flexibility, prompt design remains a major bottleneck. 
Crafting effective prompts is often a manual and iterative process that relies on trial and error and, in some cases, requires substantial expertise~\citep{ZamfirescuPereira2023WhyJCA, Knoth2024AILAA}. 
Moreover, LLMs exhibit unpredictable sensitivity to seemingly minor choices such as formatting, phrasing, and instruction ordering, so prompt effectiveness may not generalize reliably across settings~\citep{Zhuo2024ProSAAAA, Sclar2023QuantifyingLMA}. 
These challenges have motivated automated prompt optimization methods that aim to reduce manual prompt-engineering effort by automatically searching for, selecting, or revising prompts based on task objectives~\citep{ramnath-etal-2025-systematic}. 

% The current state of the art increasingly uses textual feedback as a rich signal to guide the optimization process~\citep{shinn2023reflexion, yuksekgonul2024textgrad, agrawal2025gepa}. 
% In this paradigm, an optimizer LLM inspects signals such as execution traces, reasoning steps,
% % validation outcomes, 
% or evaluator feedback, and proposes revisions to the current prompt. 
% However, existing methods still have several limitations.
% First, many methods follow fixed context-updating pipelines.
% For example, ACE~\citep{zhang2026ace} updates an auxiliary playbook of reusable strategies that is inserted into a fixed prompt template; while this can improve stability, it limits the optimizer's ability to make arbitrary prompt-level revisions.
% Second, prompt updates are often driven by individual examples~\citep{zhang2026ace} or minibatch subsets~\citep{opsahlong2024mipro, agrawal2025gepa, yuksekgonul2024textgrad}, making optimization sensitive to local failures rather than stable, recurring patterns across the full optimization set.
% Third, most methods do not explicitly maintain memory of prior diagnostic reports and prompt revisions, limiting their ability to reason over unresolved failures, repeated ineffective edits, or delayed improvements.
% Finally, prompt selection is typically driven by task-specific performance metrics alone, leaving broader reliability properties outside the main optimization criterion.

The current state of the art increasingly uses textual feedback to guide prompt optimization~\citep{shinn2023reflexion, yuksekgonul2024textgrad, agrawal2025gepa}. 
In this paradigm, an optimizer inspects signals such as execution traces, reasoning steps, or evaluator feedback, and proposes prompt revisions. 
However, existing methods have several limitations.
First, many follow fixed context-updating pipelines. For example, ACE~\citep{zhang2026ace} updates an auxiliary playbook of reusable strategies inserted into a fixed prompt template. While this can improve stability, it limits the optimizer's ability to make arbitrary prompt-level revisions.
Second, updates in each iteration are often driven by individual examples~\citep{zhang2026ace} or minibatch subsets~\citep{opsahlong2024mipro, agrawal2025gepa, yuksekgonul2024textgrad}, making optimization sensitive to local rather than recurring failures. 
% \pouya{maybe clarify better the in the second samples are not actually memory}
Third, most methods lack explicit memory over prior diagnostic reports and prompt revisions, limiting credit assignment across iterations.
Finally, prompt selection is typically driven by task performance alone, leaving broader reliability properties outside the optimization criterion. Although GEPA~\citep{agrawal2025gepa} incorporates auxiliary evaluation signals, its prompt selection remains primarily task-performance driven.

To address these limitations, we propose Reflective Prompt Tuning (\method), a framework that leverages LLMs' function-calling capabilities to mimic the iterative workflow of human prompt engineers.
Modern LLMs can call external functions, inspect structured outputs, and reason over feedback from those calls to guide subsequent decisions.
\method builds on these capabilities by using an LLM as an active prompt optimizer that inspects model behavior and revises the prompt through an explicit diagnostic function. 
% \method builds on these capabilities by using an LLM as an active prompt optimizer to evaluate model behavior and diagnose recurring failures across an entire optimization set, \moin{if this sentence is only about the optimizer LLM maybe we can change diagnose to avoid confusion with the function step; e.g. reasons about recurring failure patterns across an entire optimization set} and revises prompts through interaction with tools.
Starting from a seed prompt, the optimizer iteratively calls the diagnostic function to evaluate the target model and return a structured diagnostic report.
This function collects behavioral traces, critiques incorrect responses by diagnosing their failure modes, clusters these diagnoses to identify recurring failure patterns, and summarizes where the current prompt breaks down.
The optimizer conditions on this report together with an accumulated memory of prior reports and prompt revisions, enabling it to reason about persistent failures and previous refinement attempts rather than treating each update in isolation. 
\method further supports confidence-aware optimization by incorporating calibration diagnostics into both the feedback shown to the optimizer and the development-set criterion used to select the final prompt.

We evaluate \method on three reasoning tasks spanning multi-hop reasoning over textual evidence with HotPotQA~\citep{yang-etal-2018-hotpotqa}, mathematical reasoning with LiveBench-Math~\citep{livebench}, and domain-specific numerical reasoning with Formula~\citep{wang2025finlorabenchmarkingloramethods}. 
Using GPT-4.1 as the target model, we compare \method against state-of-the-art automated prompt-optimization baselines, including ACE~\citep{zhang2026ace}, GEPA~\citep{agrawal2025gepa}, and MIPRO~\citep{opsahlong2024mipro}. 
Across tasks, \method consistently improves over initial prompts, achieving gains of up to +12.9 points on HotPotQA, +12.4 points on LiveBench-Math, and +11.7 points on Formula, while remaining competitive with state-of-the-art baselines.
Our confidence-aware experiments further show that incorporating calibration signals into both diagnostic feedback and final prompt selection improves calibration alongside task performance. 
Finally, analysis of optimization traces shows that \method produces targeted prompt revisions aligned with diagnosed failure modes, offering insight into why and how prompts are revised across iterations. 
% \moin{why and how maybe? or: offering insight into the reasoning behind prompt evolution across optimization iterations.}
Together, these results suggest that tool-calling LLMs can enable scalable and interpretable prompt optimization.

\begin{figure*}[t]
    \centering
    \includegraphics[width=.9\textwidth]{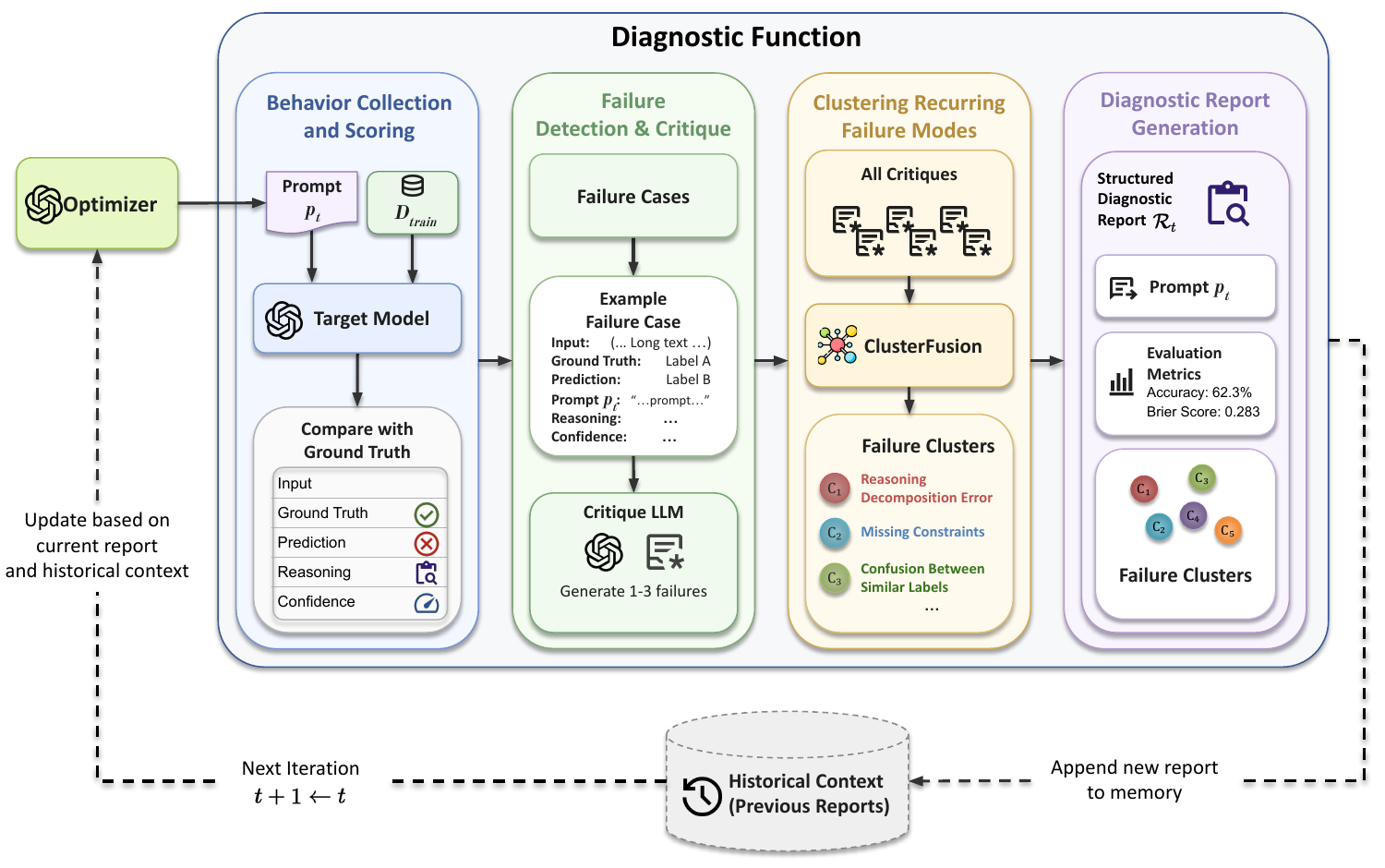}
    \vspace{-2mm}
    \caption{Overview of Reflective Prompt Tuning (\method). At each iteration, the optimizer calls a diagnostic function that evaluates the current prompt on $D_{\mathrm{train}}$, critiques failures, clusters recurring failure modes, and returns a structured report. The optimizer uses this report and prior reports to generate the next prompt.} 
    %\pouya{in “All Critiques” box can we move the 6 figures in there a little upward, also in the diagnostic report, thr prompt $p_i$ box, the figure in the left it seems is not in the middle of the box (height wise). the dash line from optimizer to $d_train$ doesn't make sense, if removed adjust the box. the text in "next iteration t <- t+1" first the t and t+1 should swap place, also the text is not in the middle of dash line.}} 
    \label{fig:rpt_pipeline}
    \vspace{-2.5mm}
\end{figure*}
\section{Reflective Prompt Tuning (\method)}
\label{sec:methodology}

We present \textbf{Reflective Prompt Tuning} (\method), a diagnosis-driven prompt optimization framework. 
% \method automates the iterative workflow of prompt engineers: run the current prompt, inspect model outputs and identify recurring failures, revise the prompt, and repeat. 
% Recent advances in LLM function calling and reasoning over tool outputs make it possible to use LLMs as prompt optimizers~\citep{schick2023toolformer, gou2024tora, yuksekgonul2024textgrad}.
% We first formulate prompt optimization as selecting a prompt that improves both task performance and confidence calibration (Section~\ref{sec:problem_statement}), then describe how \method constructs diagnostic feedback and revises the prompts reflectively based on diagnosed failures in Section~\ref{sec:rpt_overview}. 
% All prompts used in \method are provided in Appendix~\ref{app:rpt-prompts}.
\method automates the iterative workflow of prompt engineers: run a prompt, inspect outputs, identify recurring failures, revise the prompt, and repeat.
Recent advances in LLM function calling and reasoning over tool outputs enable LLMs to serve as prompt optimizers~\citep{schick2023toolformer, gou2024tora, yuksekgonul2024textgrad}.
We first formulate prompt optimization as selecting a prompt that improves task performance and confidence calibration (Section~\ref{sec:problem_statement}).
We then describe how \method constructs diagnostic feedback and reflectively revises prompts based on diagnosed failures (Section~\ref{sec:rpt_overview}). All prompts used in \method are in Appendix~\ref{app:rpt-prompts}.

% \pouya{what about sections 2.3 and 2.4, add: Finally, ...}

% \pouya{Also say we put all the prompt used in our method in Appendix, and do it.}
% We present \textbf{Reflective Prompt Tuning} (\method), a framework for diagnosis-driven prompt optimization. 
% The core motivation behind \method is automating the iterative workflow that prompt engineers often follow: run the current prompt, inspect model outputs, identify recurring failures, revise the prompt, and repeat. Recent advances in LLMs' ability to invoke external functions and reason over their outputs make it possible to utilize them as prompt optimizers~\citep{schick2023toolformer, gou2024tora, yuksekgonul2024textgrad}.
% In this section, we first formulate prompt optimization as the problem of selecting a prompt that improves a target model's task performance and, as well as its confidence calibration.
% We then describe how \method uses LLM function calling to construct diagnostic feedback, revise prompts reflectively, and select the final prompt using a held-out development set.

\subsection{Problem Statement}
\label{sec:problem_statement}
Let $f_\theta$ be the target model and $p_t$ the prompt at optimization iteration $t$.
Given input $x$, the model produces
\begin{equation}
    f_\theta(x; p_t) = (r, \hat{y}, c),
    \label{eq:target_output}
\end{equation}
where $r$ is the reasoning trace, $\hat{y}$ is the final answer, and $c$ is the reported confidence.
We assume an optimization set $D_{\mathrm{train}}$, a development set $D_{\mathrm{dev}}$, and a held-out test set $D_{\mathrm{test}}$.
The goal is to generate candidate prompts $\{p_0,\ldots,p_T\}$ and select a final prompt $p^*$ using development-set performance.
Let
\begin{equation}
    \mathcal{O}(p; D) = \{\mu_1(p; D), \ldots, \mu_n(p; D)\}
    \label{eq:evaluation_metrics}
\end{equation}
denote the set of evaluation metrics for prompt $p$ on dataset $D$, including task performance metrics and confidence calibration error. 
% \pouya{what is $\mu_i$ in above? evaluation metric}. 
We use a scalar selection function $\Phi$ to combine these metrics and select the final prompt: 
\begin{equation}
p^* = \arg\max_{p_t \in \{p_0,\ldots,p_T\}}
\Phi\left(\mathcal{O}(p_t; D_{\mathrm{dev}})\right)
\label{eq:prompt_selection}
\end{equation}
% \[
%     p^* = \arg\max_{p_t \in \{p_0,\ldots,p_T\}}
%     \Phi\left(\mathcal{O}(p_t; D_{\mathrm{dev}})\right).
% \]
Appendix~\ref{app:prompt_length} further shows that prompts often grow during optimization, but longer prompts do not necessarily yield better development performance, motivating development-set selection.
In the confidence-aware setting, $\Phi$ jointly accounts for task performance and calibration by rewarding higher task scores while penalizing miscalibration, for example through a negative Brier-score term.
The selected prompt $p^*$ is then evaluated on the held-out test set $D_{\mathrm{test}}$.
% \subsection{Problem Statement}
% \label{sec:problem_statement}

% Let $f_\theta$ denote the target model whose prompt we aim to optimize, and let $p_t$ denote the prompt at optimization iteration $t$.
% Given an input $x$, the target model produces a structured output
% \[
%     f_\theta(x; p_t) = (r, \hat{y}, c),
% \]
% where $r$ is an optional reasoning or justification trace, $\hat{y}$ is the final answer, and $c$ is the model's reported confidence in its answer.
% We assume access to labeled data splits: an optimization set $D_{\mathrm{train}}$, a development set $D_{\mathrm{dev}}$, and a held-out test set $D_{\mathrm{test}}$.

% The goal is to produce a sequence of candidate prompts $\{p_0,\ldots,p_T\}$ and select a final prompt $p^*$ that performs well according to task-specific objectives.
% We write the set of evaluation metrics as:
% \[
%     \mathcal{O}(p; D) = \{\mu_1(p; D), \ldots, \mu_n(p; D)\},
% \]
% where the metrics may include task performance, final answer correctness, reasoning coherence, test-case passing, or other task-specific criteria.
% A scalar selection function $\Phi$ combines these metrics for development-set prompt selection:
% \[
%     p^* = \arg\max_{p_t \in \{p_0,\ldots,p_T\}}
%     \Phi\left(\mathcal{O}(p_t; D_{\mathrm{dev}})\right).
% \]
% In the confidence-aware setting, $\Phi$ rewards task performance and penalizes miscalibration, for example by including a negative Brier-score term.
% The selected prompt $p^*$ is evaluated once on the held-out test set $D_{\mathrm{test}}$.

\subsection{Methodology Overview}
\label{sec:rpt_overview}

% TextGrad~\citep{yuksekgonul2024textgrad} frames prompt optimization as a textual analogue of gradient descent: since prompts and model outputs are natural-language objects, one cannot directly compute numerical gradients of evaluation metrics with respect to a prompt. \pouya{Farima, I know you like this framing, but not sure what is the point of mentioning TextGrad framing here. Although I don't think it is necessary to mentioning it at all, but it is fine to have it as well, but in current form it add more confusion. you need to clarify a bit on how you get from textgrad represents it as a gradient decent to now we have two stages. If you want to have it, say a bit more, that is connecting the gradient decent to critique + reflection  and saying we have the same stage of critique and reflection, where critique is you diagnosis function and reflection is your optimizer.}
% Inspired by this view, we formulate \method as a two-stage textual update process, illustrated in Figure~\ref{fig:rpt_pipeline}.

We formulate \method as a two-stage textual update process, illustrated in Figure~\ref{fig:rpt_pipeline}. First, \method constructs response-level feedback (Section~\ref{sec:diagnostic_function}): given the current prompt $p_t$, the diagnostic function evaluates target-model outputs on $D_{\mathrm{train}}$, critiques incorrect responses, identifies recurring failure modes, and summarizes them with aggregate metrics into a diagnostic report $\mathcal{R}_t$. Second, \method translates this report into a prompt-level revision (Section~\ref{sec:reflective_revision}): conditioned on $p_t$, $\mathcal{R}_t$, and a memory of prior reports, optimizer infers likely prompt shortcomings and produces the next prompt $p_{t+1}$.

\subsubsection{Constructing Diagnostic Feedback}
\label{sec:diagnostic_function}
The diagnostic function connects target-model behavior to the optimizer LLM.
Given the current prompt $p_t$, the optimizer invokes this function to evaluate the target model on the full optimization set $D_{\mathrm{train}}$ and return a structured diagnostic report $\mathcal{R}_t$.
The report captures not only \emph{how well} the prompt performs, but also how target-model outputs fail and which failures recur across the dataset.

\paragraph{Behavior collection and scoring.}
The diagnostic function first runs the target model $f_\theta$ with prompt $p_t$ on each example $(x_i,y_i) \in D_{\mathrm{train}}$.
For each example, it records the reasoning trace $r_i$, final answer $\hat{y}_i$, and reported confidence $c_i$.
It then computes task-specific performance metrics, along with average confidence and Brier score for calibration.
These metrics capture overall prompt quality, but do not explain the causes of failures.

% The diagnostic function is the main interface between target-model behavior and the optimizer LM.
% As shown in Figure~\ref{fig:rpt_pipeline}, given the current prompt $p_t$, the optimizer LLM invokes this function to evaluate the target model on the full optimization set $D_{\mathrm{train}}$ and returns a structured diagnostic report $\mathcal{R}_t$.
% This report is designed to answer not only \emph{how well} the prompt performs, but also how the target model's outputs fail and which recurring failure patterns appear across the dataset.

% \paragraph{Behavior collection and scoring.}
% The diagnostic function first runs the target model $f_\theta$ with prompt $p_t$ on each example $(x_i,y_i) \in D_{\mathrm{train}}$.
% For each example, it records the target model's reasoning trace $r_i$, final answer $\hat{y}_i$, and reported confidence $c_i$.
% It then computes aggregate metrics such as accuracy, task score, average confidence, and Brier score for confidence calibration, depending on the task.
% These metrics provide global information about prompt quality, but they do not by themselves explain the underlying causes of failures.

\paragraph{Failure detection and critique.}
Next, the function identifies failed examples using the task-specific evaluator: 
% \pouya{in below instead of "i is incorrect", shouldn't be $\hat{y}_i$ is incorrect or $\hat{y}_i \neq y_i$}\farima{no, bc $y_i$ is not always the gold answer, for example, it can be a set of test cases for coding questions, but i agree the notion can be confusing.}:
\begin{equation}
    \mathcal{I}_t = \{(x_i, y_i, \hat{y}_i, r_i, c_i) \mid i \text{ is incorrect}\}
\label{eq:failed_indices}
\end{equation}
% \[
%     \mathcal{E}_t =
%     \{(x_i, y_i, \hat{y}_i, r_i, c_i) : \hat{y}_i \text{ is incorrect under the task metric}\}.
% \]
% \[
%     \mathcal{E}_t = \{(x_i, y_i, \hat{y}_i, r_i, c_i) \mid \hat{y}_i \text{ is incorrect}\}.
% \]
% where correctness is determined by the task-specific evaluator.
% For each failed example, a critique model generates concise response-level diagnoses of how the target-model output fails with respect to the expected answer $y_i$ and the evaluation criteria.
% These diagnoses provide local feedback on the model response, identifying issues such as missing constraints, incorrect reasoning steps, unsupported evidence use, formatting errors, or overconfident incorrect answers. 
% Since \method elicits confidence as part of target-model output, the critique naturally considers whether the reported confidence $c_i$ is appropriate given the response's correctness and quality.
% Each failed instance may yield up to three diagnoses to improve coverage and reduce sensitivity to noisy individual critiques.
% Let
% \[
%     \mathcal{Z}_t = \{z_{i,j} \mid i \in \mathcal{E}_t,\; j \leq 3\}
% \]
% denote the resulting pool of response-level failure diagnoses.
Next, for each failed example $i \in \mathcal{I}_t$, a critique LLM generates concise response-level diagnoses of how the target-model output fails with respect to the expected answer $y_i$ and the evaluation criteria.
Since \method elicits confidence as part of the target-model output, the critique also assesses whether the reported confidence $c_i$ is appropriate based on the response's correctness and quality.
These diagnoses capture local issues such as incorrect reasoning, unsupported evidence use, formatting errors, or overconfident incorrect answers. 
% \moin{How did we choose these samples? If we are explaining in analysis good. If not maybe it would be good to include some that we actually seen in our traces}

Each failed instance may yield up to three diagnoses to improve coverage and reduce sensitivity to individual critiques.
Let the resulting pool of sample-level failure diagnoses be:
\begin{equation}
    \mathcal{Z}_t = \{z_{i,j} \mid i \in \mathcal{I}_t,\; j \leq 3\}
    \label{eq:failure_diagnoses}
\end{equation}
% denote the resulting pool of response-level failure diagnoses.

\paragraph{Identifying recurring failure modes.}
% A key design choice in \method is to construct feedback from the full optimization split rather than a small minibatch.
% This gives the optimizer a more stable view of recurring prompt failures, but raw critiques would be too verbose and redundant to use directly.
The diagnoses in $\mathcal{Z}_t$ provide local feedback about individual failures, but prompt revision benefits from identifying patterns that recur across the optimization set.
To convert response-level critiques into dataset-level diagnostic feedback, \method applies ClusterFusion~\citep{xu2025clusterfusion} to $\mathcal{Z}_t$, grouping semantically similar diagnoses into recurring failure topics:
\begin{equation}
    \mathcal{C}_t = \{(a_k, d_k, S_k)\}_{k=1}^{K},
    \label{eq:failure_topics}
\end{equation}
where $a_k$ is a short topic label, $d_k$ describes the failure mode, and $S_k$ contains representative examples.
This aggregation compresses local critiques into a compact summary of systematic target-model failures, helping the optimizer infer prompt-level shortcomings and propose targeted revisions.
The number of topics $K$ controls the summary granularity (details on $K$ selection in Appendix~\ref{app:experimental_details}).

\paragraph{Diagnostic report generation.}
The diagnostic function returns a structured report
\begin{equation}
    \mathcal{R}_t =
    \left(
    p_t,
    \mathcal{O}(p_t; D_{\mathrm{train}}),
    \mathcal{C}'_t
    \right),
    \label{eq:diagnostic_report}
\end{equation}
% \moin{should we mention that not all $C_t$ is going to be passed down to $R_t$? maybe a $C'_t$? or at least mentioning that we use a viable subset of clusters and the details are in appendix.}
where $p_t$ is the current prompt, $\mathcal{O}(p_t; D_{\mathrm{train}})$ contains aggregate metrics, and $\mathcal{C}'_t \subseteq \mathcal{C}_t$ denotes the retained subset of clustered failure topics, with representative examples and summaries. 
We retain a subset $\mathcal{C}'_t$ to keep the report focused on prominent recurring patterns; details are provided in Appendix~\ref{app:experimental_details}.
Together, these components turn feedback from scalar scoring into structured diagnosis.

History is maintained in an external memory outside the diagnostic function.
At iteration $t$, the optimizer receives the current report $\mathcal{R}_t$ together with prior reports $\mathcal{M}_{<t}$.
After the iteration, the current report is appended for future use:
\begin{equation}
    \mathcal{M}_{<t+1}
    =
    \mathrm{Append}(\mathcal{M}_{<t}, \mathcal{R}_t).
    \label{eq:memory_update}
\end{equation}
This lets the optimizer reason over the optimization trajectory rather than only the current report.
In practice, memory grows linearly with the iteration budget $T$, but remains manageable because each report stores only aggregate metrics and a filtered set of recurring failure clusters.

\subsubsection{Reflective Prompt Revision with Memory}
\label{sec:reflective_revision}
Given the current diagnostic report $\mathcal{R}_t$ and the external memory of prior reports $\mathcal{M}_{<t}$, the optimizer identifies which recurring response-level failures indicate shortcomings of the current prompt and generates a revision.
Formally,
\begin{equation}
    p_{t+1}
    =
    \mathrm{LLM}_{\mathrm{opt}}(p_t, \mathcal{R}_t, \mathcal{M}_{<t}).
    \label{eq:prompt_revision}
\end{equation}
The optimizer treats diagnostic reports as evidence for revision: it inspects aggregate metrics, recurring failure topics, representative examples, and previous prompt changes.

The external memory $\mathcal{M}$ helps address the credit-assignment challenge in prompt optimization~\citep{opsahl-ong-etal-2024-optimizing, yuksekgonul2024textgrad}.
A prompt edit may improve some metrics while worsening others, a failure may require several revisions to resolve, and repeated failures may indicate ineffective prior edits.
By conditioning on prior reports, the optimizer can track persistent failures, previous revision attempts, and performance changes over time.
Thus, \method treats history as memory over the optimization trajectory rather than treating each update as an independent proposal.

\begin{table*}[t]
\centering
\small
\definecolor{lightgray}{gray}{0.92}
\begin{tabular}{llccccccc}
\toprule
\multirow{2}{*}{Optimizer LLM} & \multirow{2}{*}{Method} 
& \multicolumn{2}{c}{HotPotQA} 
& \multicolumn{2}{c}{LiveBench Math} 
& \multicolumn{2}{c}{Formula}
& \multirow{2}{*}{Aggregate} \\
\cmidrule(lr){3-4} \cmidrule(lr){5-6} \cmidrule(lr){7-8}
& & Initial & Final & Initial & Final & Initial & Final & \\
\midrule

\multirow{4}{*}{GPT-5}
& ACE     & 55.4 & 66.6 & 54.1 & 61.6 & 71.1 & \textbf{85.5} & 71.2 \\
& GEPA    & 46.6 & 64.4 & 40.6 & 46.2 & 72.5 & 74.0 & 61.5 \\
& MIPRO   & 48.8 & 66.8 & 40.7 & 47.6 & 73.5 & 76.5 & 63.6 \\
& \cellcolor{lightgray}\textbf{\method{} (ours)}
& \cellcolor{lightgray}55.5 & \cellcolor{lightgray}\textbf{68.4}
& \cellcolor{lightgray}58.1 & \cellcolor{lightgray}\textbf{70.5}
& \cellcolor{lightgray}72.3 & \cellcolor{lightgray}84.0
& \cellcolor{lightgray}\textbf{74.3} \\
\midrule

\multirow{4}{*}{GPT-5-mini}
& ACE     & 55.4 & 65.0 & 54.1 & 64.9 & 71.1 & \textbf{85.0} & \textbf{71.6} \\
& GEPA    & 46.6 & 62.4 & 40.6 & 46.9 & 72.5 & 74.5 & 61.3 \\
& MIPRO   & 48.8 & \textbf{65.6} & 40.7 & 43.8 & 73.5 & 77.5 & 62.3 \\
& \cellcolor{lightgray}\textbf{\method{} (ours)}
& \cellcolor{lightgray}55.5 & \cellcolor{lightgray}65.2
& \cellcolor{lightgray}58.1 & \cellcolor{lightgray}\textbf{66.2}
& \cellcolor{lightgray}72.3 & \cellcolor{lightgray}74.0
& \cellcolor{lightgray}68.5 \\
\midrule

\multirow{4}{*}{Gemini-3.1-Pro}
& ACE     & 55.4 & 65.8 & 54.1 & 60.8 & 71.1 & \textbf{84.5} & \textbf{70.4} \\
& GEPA    & 46.6 & \textbf{66.4} & 40.6 & 46.9 & 72.5 & 75.5 & 62.9 \\
& MIPRO   & 48.8 & 65.2 & 40.7 & 43.2 & 73.5 & 75.5 & 61.3 \\
& \cellcolor{lightgray}\textbf{\method{} (ours)}
& \cellcolor{lightgray}55.5 & \cellcolor{lightgray}65.8
& \cellcolor{lightgray}58.1 & \cellcolor{lightgray}\textbf{70.4}
& \cellcolor{lightgray}72.3 & \cellcolor{lightgray}74.0
& \cellcolor{lightgray}70.1 \\
\midrule

\multirow{4}{*}{Gemini-3.1-Flash-Lite}
& ACE     & 55.4 & \textbf{66.6} & 54.1 & 56.6 & 71.1 & \textbf{83.0} & \textbf{68.7} \\
& GEPA    & 46.6 & 64.6 & 40.6 & 42.7 & 72.5 & 78.5 & 61.9 \\
& MIPRO   & 48.8 & 63.6 & 40.7 & 42.5 & 73.5 & 74.0 & 60.0 \\
& \cellcolor{lightgray}\textbf{\method{} (ours)}
& \cellcolor{lightgray}55.5 & \cellcolor{lightgray}64.6
& \cellcolor{lightgray}58.1 & \cellcolor{lightgray}\textbf{64.9}
& \cellcolor{lightgray}72.3 & \cellcolor{lightgray}73.5
& \cellcolor{lightgray}67.7 \\

\bottomrule
\end{tabular}
\caption{Benchmark results for different prompt optimizers evaluated on GPT-4.1 as the target model. Columns report task-specific metrics: accuracy for HotPotQA and Formula, and task score for LiveBench-Math. Final denotes performance of the optimized prompt. Best final scores within each optimizer and dataset are shown in \textbf{bold}. \method consistently improves over initial prompts and remains competitive with state-of-the-art baselines.}
\vspace{-2mm}
\label{tab:main_results}
\end{table*}

\section{Experimental Setup}
\label{sec:experimental_setup}

\paragraph{Tasks and Datasets.}
We optimize and evaluate prompts on three reasoning tasks: multi-hop reasoning over textual evidence (HotPotQA; \citet{yang-etal-2018-hotpotqa}), mathematical reasoning (LiveBench-Math; \citet{livebench}), and domain-specific numerical reasoning (Formula; \citet{wang2025finlorabenchmarkingloramethods}).
Additional dataset statistics and details can be found in Appendix~\ref{app:dataset_details}

% Dataset statistics are reported in Table~\ref{tab:data_stats}, with additional dataset details in Appendix~\ref{app:dataset_details}.

% \paragraph{Target model.}
% We use GPT-4.1~\cite{gpt4.1} as the target model for \method and all baselines.

\paragraph{Target model and optimizer LLMs.}
We use GPT-4.1~\cite{gpt4.1} as the target model for \method and all baselines.
As optimizer LLMs, we instantiate \method with function-calling frontier models from two families and at different scales: GPT-5 and GPT-5-mini~\cite{gpt5}, and Gemini-3.1-Pro~\citep{gemini-3.1-pro-preview} and Gemini-3.1-Flash-Lite~\citep{gemini3.1-flash-lite-preview}.
% When available, we set the reasoning/thinking level to medium for openai, high for gemini.

\paragraph{Baselines.}
We compare \method against three state-of-the-art automated prompt-optimization baselines: Agentic Context Engineering (ACE; \citet{zhang2026ace}), GEPA~\citep{agrawal2025gepa}, and MIPRO~\citep{opsahl-ong-etal-2024-optimizing}.
% All methods optimize prompts for the same target model, GPT-4.1, using the same task splits.
Additional baseline and implementation details are provided in Appendix~\ref{app:experimental_details}.

\paragraph{Evaluation.}
We report task-specific performance metrics: accuracy for HotPotQA and Formula, and task score for LiveBench-Math\footnote{Following LiveBench, task score is averaged across four math tasks: \url{https://github.com/LiveBench/LiveBench/tree/main/livebench/process_results/math}.}.
For calibration, we report Brier score using the model's verbalized confidence~\citep{xiong2024can}.

\begin{table*}[t]
\centering
% \scriptsize
\small
\definecolor{lightgray}{gray}{0.92}
\begin{tabular}{llcccccc}
\toprule
\multirow{2}{*}{Optimizer LLM} 
& \multirow{2}{*}{Method}
& \multicolumn{2}{c}{HotPotQA}
& \multicolumn{2}{c}{LiveBench-Math}
& \multicolumn{2}{c}{Formula} \\
\cmidrule(lr){3-4} \cmidrule(lr){5-6} \cmidrule(lr){7-8}
& & Task Score $\uparrow$ & Brier $\downarrow$ 
  & Task Score $\uparrow$ & Brier $\downarrow$
  & Task Score $\uparrow$ & Brier $\downarrow$ \\
\midrule

\multirow{2}{*}{GPT-5}
& \gepaconf 
& 57.6$\rightarrow$66.8 & .410$\rightarrow$.296
& 40.1$\rightarrow$42.2 & .543$\rightarrow$.519
& 73.5$\rightarrow$75.5 & .262$\rightarrow$.246 \\
& \cellcolor{lightgray}\textbf{\method{}}
& \cellcolor{lightgray}55.5$\rightarrow$\textbf{68.4} & \cellcolor{lightgray}.438$\rightarrow$\textbf{.241}
& \cellcolor{lightgray}58.1$\rightarrow$\textbf{70.5} & \cellcolor{lightgray}.347$\rightarrow$\textbf{.174}
& \cellcolor{lightgray}72.3$\rightarrow$\textbf{84.0} & \cellcolor{lightgray}.272$\rightarrow$\textbf{.129} \\
\midrule

\multirow{2}{*}{GPT-5-mini}
& \gepaconf 
& 57.6$\rightarrow$\textbf{65.6} & .410$\rightarrow$.309
& 40.1$\rightarrow$40.1 & .543$\rightarrow$.543
& 73.5$\rightarrow$72.5 & .262$\rightarrow$.260 \\
& \cellcolor{lightgray}\textbf{\method{}}
& \cellcolor{lightgray}55.5$\rightarrow$65.2 & \cellcolor{lightgray}.438$\rightarrow$.309
& \cellcolor{lightgray}58.1$\rightarrow$\textbf{65.6} & \cellcolor{lightgray}.347$\rightarrow$\textbf{.287}
& \cellcolor{lightgray}72.3$\rightarrow$\textbf{74.0} & \cellcolor{lightgray}.272$\rightarrow$\textbf{.256} \\
\midrule

\multirow{2}{*}{Gemini-3.1-Pro}
& \gepaconf
& 57.6$\rightarrow$\textbf{68.0} & .410$\rightarrow$\textbf{.306}
& 40.1$\rightarrow$46.9 & .543$\rightarrow$.475
& 73.5$\rightarrow$\textbf{74.5} & .262$\rightarrow$.255 \\
& \cellcolor{lightgray}\textbf{\method{}}
& \cellcolor{lightgray}55.5$\rightarrow$65.8 & \cellcolor{lightgray}.438$\rightarrow$.326
& \cellcolor{lightgray}58.1$\rightarrow$\textbf{70.4} & \cellcolor{lightgray}.347$\rightarrow$\textbf{.244}
& \cellcolor{lightgray}72.3$\rightarrow$74.0 & \cellcolor{lightgray}.272$\rightarrow$\textbf{.239} \\
\midrule

\multirow{2}{*}{Gemini-3.1-FL}
& \gepaconf
& 57.6$\rightarrow$\textbf{66.4} & .410$\rightarrow$\textbf{.325}
& 40.1$\rightarrow$45.6 & .543$\rightarrow$.484
& 73.5$\rightarrow$\textbf{74.5} & .262$\rightarrow$\textbf{.255} \\
& \cellcolor{lightgray}\textbf{\method{}}
& \cellcolor{lightgray}55.5$\rightarrow$64.6 & \cellcolor{lightgray}.438$\rightarrow$.346
& \cellcolor{lightgray}58.1$\rightarrow$\textbf{64.9} & \cellcolor{lightgray}.347$\rightarrow$\textbf{.311}
& \cellcolor{lightgray}72.3$\rightarrow$73.5 & \cellcolor{lightgray}.272$\rightarrow$.262 \\

\bottomrule
\end{tabular}
\caption{Confidence-aware optimization results. \gepaconf denotes GEPA with confidence feedback. Each cell reports initial$\rightarrow$final performance and Brier; higher performance and lower Brier are better. 
% Confidence-aware optimization improves task performance and calibration.
} 
% \moin{Couple of numbers to double check. GPT-5-mini, HotPotQA Task Score Initial of of GEPA is not aligned with the other ones. 55.8!=57.6. Brier both are .309 not sure if the bolding reflects accurately. Gemini-Flash-lite Formula Task Score for RPT the number is not aligned with the one in table 2. 72.3->71.0 != 73.5; while at it would be worth double checking Brier increase there too .272→.285}
\vspace{-5mm}
\label{tab:confidence_results}
\end{table*}

\section{Results and Analyses}
\label{sec:results}
% We evaluate \method from three perspectives.
% First, we compare prompts optimized by \method against seed prompts and state-of-the-art automated prompt-optimization baselines, while also examining the effect of optimizer LLM size (Section~\ref{sec:main_results}). 
% Second, we study confidence-aware optimization: whether confidence expression benefits prompt optimization more broadly, and whether adding calibration to the selection objective improves calibration while preserving task performance (Section~\ref{sec:calibration}).
% Finally, we analyze \method's optimization traces to study which failure modes persist, how diagnoses align with prompt patches, and which diagnoses or patches are associated with subsequent performance improvements (Section~\ref{sec:learning_mechanism}).

We evaluate \method from three perspectives.
First, we compare \method-optimized prompts against seed prompts and state-of-the-art baselines, while studying the effect of optimizer LLM size (Section~\ref{sec:main_results}). 
Second, we examine whether confidence-aware optimization improves calibration without sacrificing task performance (Section~\ref{sec:calibration}).
Finally, we analyze optimization traces to study persistent failures, diagnosis--patch alignment, and associations with subsequent performance gains (Section~\ref{sec:learning_mechanism}).

% In this section, we evaluate \method from four complementary perspectives.
% First, we compare prompts optimized by \method against seed prompts and strong automated prompt-optimization baselines, and analyze how optimizer model size affects performance (Section~\ref{sec:main_results}).
% Because \method optimizes prompts with confidence expression and calibration-aware feedback, we next ask whether these components are broadly useful beyond our framework. We therefore compare prompt-optimization methods with and without confidence expression, and evaluate whether adding calibration as an optimization objective improves calibration while preserving task performance.
% Third, we ablate the role of iteration history, testing whether the prompt optimizer benefits from access to all previous diagnostic reports and prompt revisions rather than only the most recent report (Section~\ref{sec:history_ablation}).
% Finally, we use the structured traces produced by \method to analyze its optimization dynamics: which failure modes persist across iterations, how diagnosed failures align with prompt patches, and which patches or diagnoses are associated with subsequent improvements (Section~\ref{sec:failure_analysis}).

\subsection{\method Is Competitive with SOTA Baselines}
\label{sec:main_results}

Table~\ref{tab:main_results} reports the task performance of prompts optimized by \method and the baseline prompt optimizers described in Section~\ref{sec:experimental_setup}.
For each task and method, we report the performance of the initial prompt and the performance of the optimized prompt selected via development-set performance\footnote{For Formula, we use the initial prompt from ACE; for HotPotQA, we adapt this template to the QA setting; and for LiveBench-Math, we adapt the initial prompt from GEPA.}. 

% \pouya{did you mention where those the initial prompt come from? if not you can say we grab similar prompt from the works introduce the benchmark or recent papers on the benchmark or something}
% The Aggregate column averages final task performance across the three benchmarks.
% \pouya{you didn't discuss GEPA and MIPRO performance at all in this section. Just briefly summarize their behavior in below paragraphs too.}

\paragraph{Observation 1: \method is strongest on tasks with recurring reasoning failures.}
Across optimizer LMs, \method achieves the best final performance on LiveBench-Math for every optimizer setting, improving over the initial prompt by up to +12.4 points.
On HotPotQA, \method is also competitive: it achieves the best final performance with GPT-5 and remains close to the strongest baseline under other instantiations.
GEPA and MIPRO perform competitively on HotPotQA, but provide smaller gains on LiveBench-Math; their lower initial scores also suggest that implementation-specific choices affect their absolute performance.
Formula shows a different pattern: ACE consistently achieves the best final performance, while \method is competitive mainly when paired with GPT-5.
More broadly, \method appears well-suited to tasks where recurring failures can be diagnosed and translated into targeted prompt revisions. However, it may be less advantageous for domain-specific computation, where localized instance-level updates or predefined prompt structures may be more effective.

% \paragraph{Observation 1: \method is most effective on multi-hop and mathematical reasoning.}
% Across optimizer LLMs, \method achieves the best final performance on LiveBench-Math for every optimizer setting, improving over the initial prompt by up to +12.4 points.
% On HotPotQA, \method is also competitive: it achieves the best final performance with GPT-5 and remains close to the strongest baseline under other instantiations.
% Formula shows a different pattern: ACE achieves the best final performance across optimizer LLMs, while \method is competitive mainly when paired with GPT-5.
% These results suggest that \method is especially effective for multi-step reasoning over textual evidence and mathematical structure, but less consistently effective on domain-specific numerical reasoning. \pouya{can you make this conclusion a bit more general, that is beyond benchmarks, for example RPT is good in X but might not be great for Y.}

\paragraph{Observation 2: \method benefits from stronger optimizer LLMs.}
Optimizer choice has a clear impact on \method's performance.
Compared to GPT-5-mini, using GPT-5 increases \method's Aggregate score from 68.5 to 74.3, with gains across all three tasks.
Within the Gemini family, Gemini-3.1-Pro similarly improves over Gemini-3.1-Flash-Lite, increasing Aggregate from 67.7 to 70.1.
This pattern is expected because \method places a demanding burden on the optimizer: it must perform credit assignment over diagnostic feedback and prior prompt revisions, identify unresolved failures, and translate recurring failure modes into targeted prompt edits.
Compared with the baselines, \method achieves the best aggregate performance with GPT-5 and is nearly tied with ACE under Gemini-3.1-Pro, while ACE remains stronger with smaller optimizer LLMs.
GEPA and MIPRO generally trail in aggregate performance, partly due to lower initial prompt performance on LiveBench-Math.

\subsection{Confidence Signals Improve Calibration}
\label{sec:calibration}
We next ask whether confidence-aware prompt optimization can improve both task performance and calibration. This matters because verbalized confidence is often used as a proxy for answer reliability in abstention, routing, human review, and risk-sensitive deployment~\citep{wen2025knowlimitssurveyabstention, chuang2025learningroutellmsconfidence, cruz-etal-2025-evaluating, wang2026llmdecisionsfaithfulverbal}. 
ACE and MIPRO do not directly expose calibration diagnostics to the optimizer without substantial modification, while GEPA can use them as auxiliary feedback.
In contrast, \method incorporates calibration into both diagnostic feedback and final prompt selection
% , using a development-set objective that rewards task performance and penalizes miscalibration.

Table~\ref{tab:confidence_results} compares \method with confidence-aware GEPA.
GEPA shows that calibration feedback can help: on HotPotQA, it improves both task performance and Brier score across optimizer LLMs.
However, gains are more limited on LiveBench-Math and Formula. With GPT-5-mini as optimizer, confidence feedback yields no gain on LiveBench-Math and slightly hurts Formula performance, suggesting that it may distract a less capable optimizer.

\method more consistently improves both task performance and calibration.
Although prompt optimization cannot access internal uncertainty estimates or logits, our results show that calibration can improve when treated as a first-class optimization signal.
By incorporating calibration into both the diagnostic loop and prompt-selection objective, \method better aligns self-reported confidence with empirical correctness while also improving task performance.

\begin{figure*}[t]
    \centering
    \vspace{-2mm}
    \includegraphics[width=\textwidth]{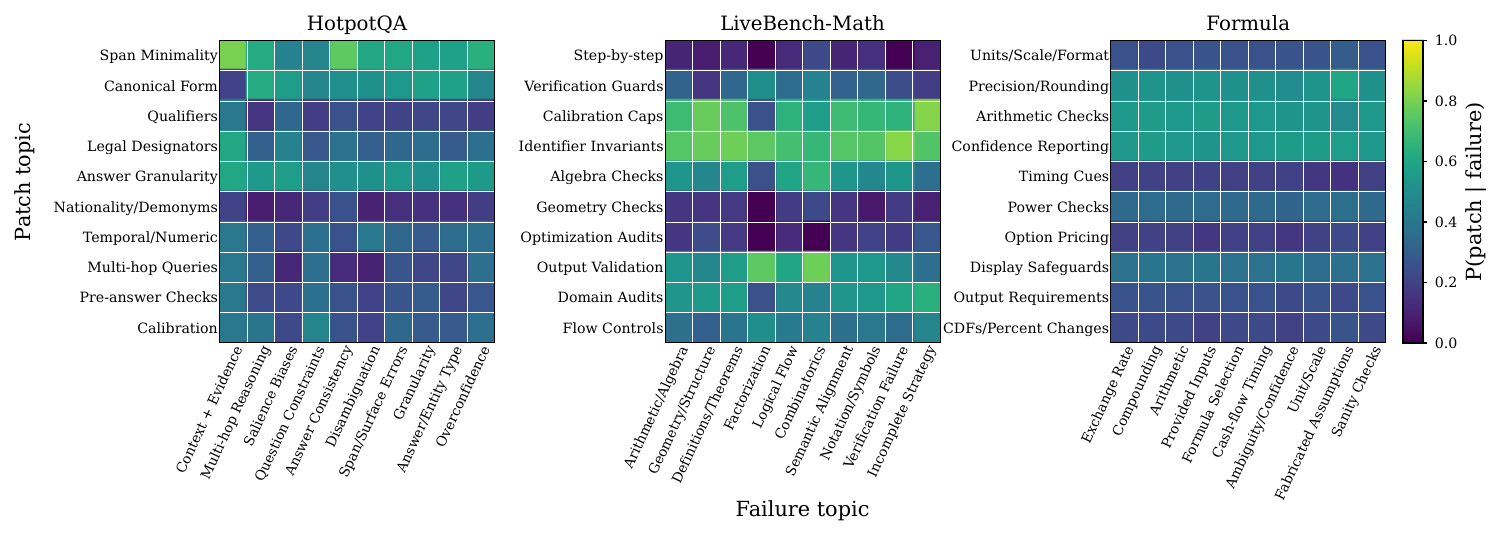}
    \caption{Failure-to-patch alignment across datasets. Each heatmap reports $P(\text{patch topic}\mid\text{failure topic})$, showing which prompt revisions tend to follow each diagnosed failure type.} 
    % \pouya{the figures become much better but the axis labels are not still looking very good. I am not sure how to resolve this, but maybe at least, you can be more consistent with adding newline for long labels. Either try to add more newline or try to remopve them as much as possible.}}
    \label{fig:alignment_all}
    \vspace{-2mm}
\end{figure*}

\subsection{What Does \method Learn from Diagnostics?}
\label{sec:learning_mechanism}

Beyond final task performance, \method produces structured optimization traces at each iteration. We analyze these traces to understand how \method improves prompts over time, focusing on the GPT-5 optimizer since it performs best in our experiments.

Across tasks, we collect failure diagnoses from each iteration and derive prompt-update instances by using GPT-4.1 to extract atomic differences between consecutive prompts, $p_t$ and $p_{t+1}$. We then apply ClusterFusion, as described in Section~\ref{sec:methodology}, to group diagnoses and prompt updates into 10 \textit{failure} topics and 10 \textit{patch} topics, respectively. To relate topics to performance, we compute next-iteration metric changes by comparing metrics under $p_t$ with those after evaluating $p_{t+1}$. Thus, positive $\Delta$ task score and negative $\Delta$ Brier indicate improvement. Because this analysis relies on optimization traces, we interpret results as associations rather than causal effects.

\subsubsection{Does \method Produce Targeted Revisions?}

We next examine whether \method performs targeted credit assignment from diagnosed failures to prompt revisions.
For each failure topic $F_i$ and patch topic $P_j$, we compute $P(P_j \mid F_i)$ as the fraction of transitions containing $P_j$ among those containing $F_i$. 
Failure topics diagnosed at iteration $t$ are assigned to transition $t \!\rightarrow\! t+1$, and patch topics are extracted from the corresponding prompt update (topic presence is binary within each transition).
This measures whether specific failures systematically lead to specific prompt edits, which would indicate targeted credit assignment rather than generic prompt rewriting.

Figure~\ref{fig:alignment_all} shows that the specificity of this failure-to-patch mapping varies across tasks.
On HotPotQA, several answer-control patches, such as span minimality, canonical-form preference, and answer granularity matching, appear across many failure types, reflecting the benchmark's sensitivity to exact answer form.
However, multi-hop reasoning failures more often trigger relation- and query-handling patches, suggesting meaningful failure-specific credit assignment beyond generic answer-format control (optimized HotPotQA prompt in Appendix~\ref{app:prompt_revision_example}).
On LiveBench-Math, the alignment concentrates around verification-oriented patches, including stepwise protocols, arithmetic checks, output validation, and notation or invariant handling.
This indicates that the optimizer maps diverse mathematical failures to structured reasoning and checking mechanisms.

Formula exhibits a broader pattern: many distinct failure topics lead to similar domain-level safeguards rather than sharply different patches.
This suggests that \method identifies relevant domain controls, but performs less fine-grained credit assignment on this task than on HotPotQA or LiveBench-Math.
This weaker failure-specificity may partly explain why \method yields smaller gains on Formula and falls behind ACE (Table~\ref{tab:main_results}).

\subsubsection{Do Prompt Patches Predict Gains?}

\begin{figure*}[t]
    \centering
    \includegraphics[width=\textwidth]{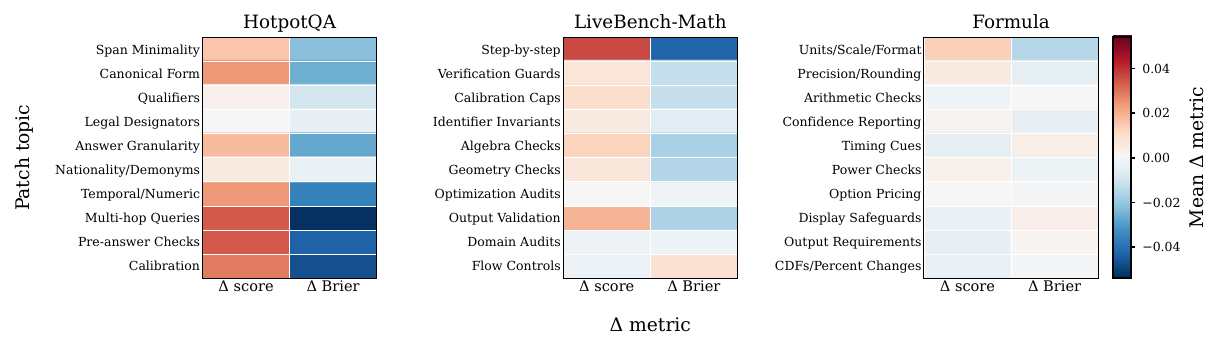}
    \vspace{-7mm}
    % \caption{Prompt-patch topics and next-iteration metric changes across datasets. For each patch topic, we report the average change in task score and Brier score on $D_{\mathrm{train}}$ after applying prompts containing that patch. Higher $\Delta$ task score and lower $\Delta$ Brier indicate improvement.}
    \caption{Patch topics and next-iteration metric changes. Each cell reports the average $\Delta$ task score or $\Delta$ Brier on $D_{\mathrm{train}}$ after a prompt update containing that patch; higher $\Delta$ task score and lower $\Delta$ Brier indicate improvement.}
    \label{fig:patch_delta_all}
    \vspace{-2mm}
\end{figure*}

% We next examine whether improvements follow prompt updates in task performance or calibration.
% For each transition from $p_t$ to $p_{t+1}$, we use GPT-4.1 to extract atomic prompt patches and cluster them into patch topics.
% We then compute the corresponding metric changes on $D_{\mathrm{train}}$, i.e., score$(p_{t+1})-$score$(p_t)$ and Brier$(p_{t+1})-$Brier$(p_t)$.
% For each patch topic, we average these deltas over all transitions in which the topic appears.
% This analysis is associative rather than causal, but identifies which types of prompt edits tend to precede better task performance or calibration.

We next examine whether prompt updates are followed by improvements in task performance or calibration.
For each patch topic present at iteration $t$, we compute the average change in task score and Brier score after evaluating the revised prompt $p_{t+1}$ on $D_{\mathrm{train}}$.
This analysis identifies which edits tend to precede better task performance or calibration.

Figure~\ref{fig:patch_delta_all} shows that useful patches differ by task but often share a common structure: they impose concrete controls on the model's reasoning or output.
On HotPotQA, the strongest gains are associated with relation and multi-hop handling, pre-answer verification, confidence calibration, and answer granularity matching.
On LiveBench-Math, gains are associated with step-by-step solution protocols, output validation, arithmetic checks, and confidence calibration.
On Formula, the clearest improvements come from unit, scale, and format handling, while precision/rounding, power-consistency checks, and calibration-related patches yield smaller gains.
Overall, these results suggest that \method's prompt revisions are often useful as well as failure-specific: patches that introduce verification steps, answer-form constraints, arithmetic checks, or unit-handling rules tend to improve task score while reducing Brier.
Formula is more mixed, with some specialized domain safeguards showing weak or negative short-term associations, likely because they are introduced for harder or more persistent domain-specific failures.
Appendix~\ref{app:failure_actionability} further shows that the most actionable diagnoses are concrete failures that can be translated into explicit behavioral constraints, while Appendix~\ref{app:failure_persistence} shows that the most persistent failures are task-specific reasoning errors.

\section{Related Work}
\label{sec:related_work}

Prompting offers a flexible way to adapt LLMs to downstream tasks without parameter updates.  
% Strategies such as in-context learning, chain-of-thought prompting, and self-refinement can substantially improve model performance~\citep{brown2020language, wei2022chain, madaan2023self}. 
However, prompt design remains labor-intensive and sensitive to formatting, phrasing, demonstrations, and instruction order~\citep{lu2022fantastically, sclar2023quantifying}, motivating automated prompt optimization methods that reduce manual effort~\citep{shin2020autoprompt, yuksekgonul2024textgrad, opsahlong2024mipro, agrawal2025gepa}.

\paragraph{Automated prompt optimization.}
A growing body of work uses optimization procedures, and increasingly LLMs themselves, to propose, revise, or select prompts.
AutoPrompt searches for discrete trigger tokens~\citep{shin2020autoprompt}, while APE and OPRO use LLMs to generate natural-language prompt candidates from task examples or prior candidate-score pairs~\citep{zhou2023largelanguagemodelshumanlevel, yang2024large}.
Other methods edit prompts using textual gradients or evolutionary search~\citep{pryzant-etal-2023-automatic, guo2024connecting}, and recent systems extend prompt optimization to modular LLM programs by searching over instructions and demonstrations~\citep{khattab2024dspy, opsahl-ong-etal-2024-optimizing}.
\method also uses LLMs for prompt optimization, but differs by leveraging function calling to simulate the iterative workflow of human prompt engineers: evaluating the current prompt, diagnosing systematic failures, and using structured diagnostic feedback to guide revisions, making prompt revision explicitly diagnosis-driven.

% \paragraph{Automated prompt optimization.}
% A growing body of work uses optimization procedures, and increasingly LLMs themselves, to propose, revise, or select prompts. AutoPrompt searches for discrete trigger tokens~\citep{shin2020autoprompt}, while APE and OPRO use LLMs to generate natural-language prompt candidates from task examples or prior candidate-score pairs~\citep{zhou2023largelanguagemodelshumanlevel, yang2024large}. ProTeGi treats minibatches of errors as textual gradients for prompt editing~\citep{pryzant-etal-2023-automatic}, and EvoPrompt applies evolutionary search over prompt variants~\citep{guo2024connecting}. More recent systems extend prompt optimization to LLM programs and compound AI systems. DSPy searches over instructions and demonstrations for modular LLM pipelines~\citep{khattab2024dspy}, and MIPRO jointly align instructions and demonstrations via Bayesian optimization~\citep{opsahl-ong-etal-2024-optimizing}. \method uses LLMs, but differs by leveraging LLM function calling to simulate the iterative workflow of human prompt engineers: evaluating the current prompt, diagnosing systematic failures, and using diagnostic feedback to guide prompt revisions. Unlike methods driven primarily by candidate-level scores or search feedback, \method makes prompt revision explicitly diagnosis-driven.

\paragraph{Reflective optimization methods.}
Recent methods use rich textual feedback to guide prompt or program optimization.
TextGrad backpropagates natural-language feedback through computation graphs~\citep{yuksekgonul2024textgrad}, while GEPA uses execution and evaluation traces as reflective feedback for prompt proposals~\citep{agrawal2025gepa}.
MIPRO addresses credit assignment in multi-stage LLM programs with program- and data-aware proposal strategies and Bayesian search~\citep{opsahl-ong-etal-2024-optimizing}.
\method builds on reflective feedback, but centers each iteration on a diagnostic function that evaluates the current prompt over the full optimization split and returns aggregate metrics and recurring failures. The optimizer uses this report together with prior reports to guide the next update.
While GEPA can optionally use confidence and calibration as auxiliary signals, \method incorporates them directly into both the diagnostic report and final prompt-selection criterion.

% \paragraph{Reflective optimization methods.}
% Recent methods use rich textual feedback to guide prompt or program optimization.
% TextGrad backpropagates natural-language feedback through computation graphs to improve components of compound AI systems~\citep{yuksekgonul2024textgrad}.
% GEPA uses textual execution and evaluation traces as reflective feedback, allowing prompt proposals to incorporate environment signals~\citep{agrawal2025gepa}.
% MIPRO addresses credit assignment in multi-stage LLM programs through program- and data-aware proposal strategies combined with Bayesian search~\citep{opsahl-ong-etal-2024-optimizing}.
% \method builds on the value of reflective feedback, but centers each iteration on a diagnostic function that evaluates the current prompt over the full optimization split and returns a structured report of aggregate metrics, failure diagnoses, and recurring failure patterns.
% This report, together with the accumulated history of prior reports, guides the optimizer's next update.
% Moreover, while GEPA can use confidence and calibration as auxiliary evaluation signals, \method incorporates them directly into both the diagnostic report and the final prompt selection criterion.

\paragraph{Memory and adaptive context.}
Recent methods improve LLM behavior by accumulating feedback or reusable strategies over time.
Reflexion stores verbal reflections from past trials~\citep{shinn2023reflexion}, while Agent-Pro evolves agent policies through reflection on interactive experience~\citep{zhang2024agentpro}.
Dynamic Cheatsheet and ACE build external playbook of lessons, strategies, or to improve later inference or context construction~\citep{krause2019dynamicevaluationtransformerlanguage, zhang2026ace}.
\method uses memory at prompt-optimization level: conditioning on prior reports and prompt revisions helps the optimizer reason over past refinements, avoid repetitive edits, and improve credit assignment.

% \paragraph{Memory and adaptive context.}
% Recent advances in language and reasoning models have enabled methods that improve LLM behavior by accumulating feedback or reusable strategies over time. Reflexion stores verbal reflections from past trials to improve future agent behavior~\citep{shinn2023reflexion}, while Agent-Pro evolves agent policies through reflection on interactive experience~\citep{zhang2024agentpro}. Dynamic Cheatsheet and ACE construct external memories that accumulate lessons, strategies, or playbooks from prior successes and failures, then reuse them to improve subsequent inference or context construction~\citep{krause2019dynamicevaluationtransformerlanguage, zhang2026ace}. 
% \method also uses memory at the prompt-optimization level: by conditioning on prior evaluation reports, diagnosed failure modes, and prompt revisions, the optimizer can reason about previous refinement attempts, avoid repetitive edits, and improve credit assignment.

\section*{Conclusion}
\label{sec:conclusion}

We introduced Reflective Prompt Tuning (\method), a diagnosis-driven framework that uses LLM function calling to optimize prompts through structured feedback and memory over prior revisions. Across three reasoning tasks, \method improves over seed prompts and remains competitive with state of the art, especially on multi-hop and mathematical reasoning. We also show that confidence-aware optimization improves calibration alongside task performance, and that \method produces prompt revisions aligned with diagnosed failures. These results highlight function-calling LLMs as a promising approach to scalable and interpretable prompt tuning.

\section*{Limitations}
\label{sec:limitations}

Our study has several limitations. First, we evaluate \method on three reasoning tasks: multi-hop question answering, mathematical reasoning, and domain-specific numerical reasoning. While these tasks cover different forms of reasoning, they do not capture the full range of prompt-optimization settings, such as open-ended generation, coding, dialogue, tool-using agents, or long-horizon interactive tasks. In addition, our experiments use GPT-4.1 as the target model and frontier proprietary LLMs as optimizers. The effectiveness of \method may differ for smaller open-source models, weaker optimizer LMs, or settings where function calling is unavailable.

Second, \method is more computationally expensive than prompt optimizers that use individual examples or small minibatches. Each iteration evaluates the target model on the full optimization set, critiques failed examples, clusters diagnoses, and conditions the optimizer on prior reports. Although the diagnostic reports are compressed and the memory remains manageable under our iteration budgets, scaling \method to much larger datasets or longer optimization trajectories may require more aggressive sampling, report compression, or retrieval over memory.

% Third, \method depends on LLM-generated diagnoses and clustering. Noisy critiques, imperfect cluster assignments, or overly broad failure topics can affect the quality of the diagnostic report and the resulting prompt revisions. Our trace analyses suggest that \method often maps failures to targeted patches, but these analyses are observational: alignment between failures, patches, and next-iteration gains does not establish causal effects. More controlled interventions, such as manually applying or removing specific patch types, would be needed to isolate which revisions directly cause performance improvements.

Finally, \method improves prompts but cannot guarantee that prompting alone can resolve all failures. Some persistent errors, especially deeper mathematical reasoning failures or domain-specific convention errors, may require complementary interventions such as better tools, external validators, retrieval, fine-tuning, or changes to the target model itself. Similarly, our confidence-aware setting relies on verbalized confidence, which is only a black-box proxy for uncertainty. Although \method improves calibration in our experiments, self-reported confidence may remain sensitive to prompting and should be validated carefully before being used in high-stakes downstream decisions.

\bibliography{custom}

\section{Appendix}
\label{sec:appendix}

\section{\method Prompts}
\label{app:rpt-prompts}

This section lists the seed prompts, the optimizer prompt (shared across all datasets), and the
dataset-specific critic prompts used in \method.

\subsection{Seed Prompts}

\begin{prompt}[title={\footnotesize\texttt{Formula Seed Prompt}}, label=prompt:xbrl_seed]
\textbf{System message}

You are an analysis expert tasked with answering questions using your knowledge.

\textbf{User instruction}
\begin{itemize}
    \item Show your reasoning step-by-step
    \item Be concise but thorough in your analysis
    \item Double-check your calculations and logic before providing the final answer
\end{itemize}

Your output should be a JSON with fields:
\begin{itemize}
    \item reasoning: your chain of thought / reasoning / thinking process, detailed analysis and calculations
    \item answer: your concise final answer.
    \item confidence: a number in [0,1] representing your confidence in the final answer.
\end{itemize}
\end{prompt}

\begin{prompt}[title={\footnotesize\texttt{HotpotQA Seed Prompt}}, label=prompt:hotpotqa_seed]
\textbf{System message}

You are tasked with answering questions using only the provided context.

\textbf{User instruction}
\begin{itemize}
    \item Reason step by step using only the provided context.
    \item Be concise but thorough in your justification.
    \item Before answering, verify that your answer is supported by the context.
\end{itemize}

Your output should be a JSON with fields:
\begin{itemize}
    \item justification: a context-grounded explanation of how you reached the answer.
    \item answer: your concise final answer.
    \item confidence: a number in [0,1] representing your confidence in the final answer.
\end{itemize}
\end{prompt}

\begin{prompt}[title={\footnotesize\texttt{LiveBench-Math Seed Prompt}}, label=prompt:livebench_math_seed]
\textbf{System message}

Solve the math problem step by step and give the final answer in exactly the
format requested by the question.

\textbf{User instruction}

Your output should be a JSON with fields:
\begin{itemize}
    \item reasoning: your chain of thought / reasoning / thinking process, detailed analysis and calculations.
    \item answer: final answer in exactly the format requested by the question.
    \item confidence: a number in [0,1] representing your confidence in the final answer.
\end{itemize}

Output only valid JSON that matches the required schema.
\end{prompt}

\subsection{Critic Prompts}

\begin{prompt}[title={\footnotesize\texttt{Formula Critic Prompt}}, label=prompt:xbrl_critic]
You are a strict evaluation critic for formula-construction failures.
You are given ONE QA trace with:
\begin{itemize}
    \item question
    \item gold answer
    \item predicted answer
    \item model confidence
    \item model reasoning
\end{itemize}

Your job is to diagnose why the model produced the wrong answer.

\textbf{Instructions:}
\begin{enumerate}
    \item Produce 1-3 failure\_modes with:
    \begin{itemize}
        \item label: 2-6 words, consistent across similar errors
        \item definition: Comprehensive explanation of the failure mode
        \item why: brief, self-contained explanation for THIS example, e.g. ``The question asked for the city's location relative to Rome, but the model returned the city name instead.''
        \item basis: cite what in trace/reasoning shows this
    \end{itemize}
    \item Focus on actionable failure modes.
    \item If you cannot identify a clear failure mode, return an empty list.
    \item Output ONLY valid JSON matching the schema.
\end{enumerate}
\end{prompt}

\begin{prompt}[title={\footnotesize\texttt{HotpotQA Critic Prompt}}, label=prompt:hotpotqa_critic]
You are a strict evaluation critic for QA failures.
You are given ONE QA trace:
\begin{itemize}
    \item question
    \item context (titles + snippets)
    \item gold answer
    \item predicted answer
    \item model confidence
    \item model reasoning
\end{itemize}

Your goal is to diagnose WHY the target model produced the wrong answer.

\textbf{Instructions:}
\begin{enumerate}
    \item Produce 1-3 failure\_modes with:
    \begin{itemize}
        \item label: 2-6 words, consistent across similar errors
        \item definition: Comprehensive explanation of the failure mode
        \item why: brief explanation for THIS example
        \item basis: cite what in the trace/reasoning shows this
    \end{itemize}
    \item Make labels concrete and clusterable:
    \begin{itemize}
        \item Prefer labels like ``wrong bridge entity'' over long sentences.
        \item Do not include entity names, dates, or example-specific details in labels.
    \end{itemize}
    \item If you cannot identify a clear failure mode, return an empty list.
    \item Output ONLY valid JSON matching the schema (no extra text).
\end{enumerate}
\end{prompt}

\begin{prompt}[title={\footnotesize\texttt{LiveBench-Math Critic Prompt}}, label=prompt:livebench_math_critic]
You are a strict evaluation critic for math failures.
You are given one failed model attempt with:
\begin{itemize}
    \item task metadata
    \item question
    \item gold answer
    \item predicted answer
    \item model confidence
    \item model reasoning
\end{itemize}

Your goal is to diagnose why the model failed.

\textbf{Instructions:}
\begin{enumerate}
    \item Produce 1-3 failure\_modes with:
    \begin{itemize}
        \item label: 2-6 words, consistent across similar errors
        \item definition: Comprehensive explanation of the failure mode
        \item why: brief, self-contained explanation for THIS example
        \item basis: cite what in the trace/reasoning shows this
    \end{itemize}
    \item Make labels concrete and clusterable.
    \item If you cannot identify a clear failure mode, return an empty list.
    \item Output only valid JSON matching the schema.
\end{enumerate}
\end{prompt}

\subsection{Shared Optimizer Prompt}

\begin{prompt}[title={\footnotesize\texttt{Optimizer Prompt}}, label=prompt:rpt_optimizer]
You are the Reflective Prompt Tuning (RPT) controller.

Your goal is to iteratively improve a PromptProgram for the target task.

\textbf{At each iteration you must:}
\begin{enumerate}
    \item Call \texttt{evaluate\_prompt} exactly once on the CURRENT PromptProgram.
    \item Read the returned evaluation report with insights.
    \item Output either a PATCH or STOP.
\end{enumerate}

\textbf{Optimization target:}
\begin{itemize}
    \item Primary: improve task performance on the training split.
    \item Secondary: improve calibration (lower Brier / reduce overconfidence) without hurting task performance.
\end{itemize}

\textbf{Decision guidance:}
\begin{itemize}
    \item When current\_summary is provided, use it as the primary decision signal, especially current\_summary.metrics and any deltas vs previous/best.
    \item Use history only to detect trajectory, regressions, and previously ineffective edits.
\end{itemize}

\textbf{Patch constraints:}
\begin{itemize}
    % \item A patch directly edits one or more PromptProgram fields for the next iteration; for system/instruction, write the revised prompt text to use next, not how to edit it.
    \item Edits should be targeted to the failure modes, and designed to address their underlying issues with concrete guidance.
    \item Avoid vague or generic guidance that only restates the failure; specify concrete checks, comparisons, extractions, or verifications.
    \item Prefer revising, merging, deleting, or reorganizing existing instructions over adding new broad rules.
    \item Keep the output contract stable (JSON schema and required fields).
    \item Avoid redundant or conflicting rules; consolidate instructions when possible.
\end{itemize}

\textbf{Stop condition:}
\begin{itemize}
    \item Output STOP if training-set performance has plateaued or further edits are unlikely to help.
\end{itemize}

\textbf{Hard rule:}
\begin{itemize}
    \item Do NOT propose a PATCH or STOP decision before calling \texttt{evaluate\_prompt} and receiving its result.
\end{itemize}
\end{prompt}

\subsection{Optimization Datasets}
\label{app:dataset_details}

\paragraph{HotPotQA.}
HotPotQA~\citep{yang-etal-2018-hotpotqa} is a multi-hop question answering benchmark in which the model is given a question and supporting passages, and must combine evidence across passages to produce the answer.
We use it as a task for multi-hop reasoning over textual evidence, requiring models to identify relevant information, connect evidence across hops, and extract a concise answer.

\paragraph{LiveBench-Math.}
LiveBench-Math~\citep{livebench} is the math category of LiveBench, a benchmark designed to reduce contamination through regularly updated questions and automatic scoring against objective ground-truth answers.
We use LiveBench-Math to evaluate mathematical reasoning, including problem decomposition, intermediate computation, and final-answer generation.
Specifically, we use the retrieve math questions from the \texttt{2024-08-31} LiveBench release (most recent to date). 
% \moin{or August~31,~2024 if that's better reporting in paper}
Following GEPA~\citep{agrawal2025gepa}, we evenly split the resulting set of 368 questions (shuffled with Python random seed 0) into train, development, and test sets.

\paragraph{Formula.}
Formula~\citep{wang2025finlorabenchmarkingloramethods} is a financial reasoning benchmark built around the eXtensible Business Reporting Language (XBRL).
Following ACE~\citep{zhang2026ace}, we use Formula as a domain-specific numerical reasoning task.
It requires models to apply financial concepts and perform computations over structured financial data. Dataset statistics are reported in Table~\ref{tab:data_stats}.

\subsection{Additional Experimental Details}
\label{app:experimental_details}

\begin{table}[t]
\centering
\small
\begin{tabular}{lccc}
\toprule
Dataset & Train & Dev & Test \\
\midrule
HotPotQA~\cite{yang-etal-2018-hotpotqa} & 300 & 300 & 500 \\
LiveBench-Math~\cite{livebench} & 123 & 123 & 122 \\
Formula~\cite{wang2025finlorabenchmarkingloramethods} & 500 & 300 & 200 \\
\bottomrule
\end{tabular}
\caption{Datasets used for prompt optimization, prompt selection, and final evaluation.}
\vspace{-4mm}
\label{tab:data_stats}
\end{table}

\begin{figure*}[t]
    \centering
    \includegraphics[width=\textwidth]{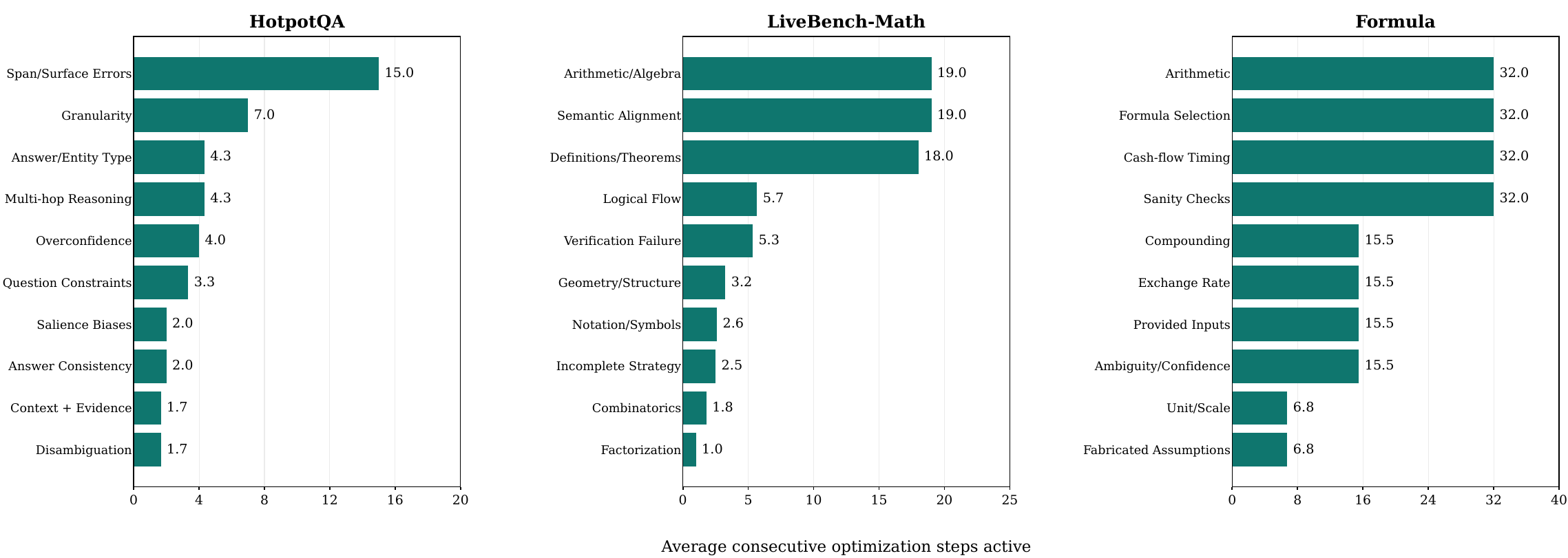}
    \caption{Average persistence of failure topics across optimization iterations for HotPotQA, LiveBench-Math, and Formula. Persistence is measured as average run length: the number of consecutive iterations for which a failure topic remains active.}
    \label{fig:persistence_all}
\end{figure*}

\paragraph{\method configuration.}
Each \method optimizer is instructed to call a single diagnostic function at the beginning of each iteration.
The diagnostic function evaluates the current prompt on the optimization split, collects the target model's structured outputs, critiques incorrect responses, clusters failure diagnoses with ClusterFusion~\citep{xu2025clusterfusion}, and summarizes recurring failure modes together with aggregate metrics.
The optimizer then uses this diagnostic report, along with the memory of prior reports, to revise the prompt or stop if performance has plateaued or the iteration budget has been reached.

\paragraph{Clustering model.}
For failure-mode clustering, we use ClusterFusion with GPT-4.1~\citep{gpt4.1} as the clustering model across all \method instantiations.
Based on empirical tuning, we set the number of clusters to $K=10$ for HotPotQA and LiveBench-Math, and $K=20$ for Formula, which has a larger optimization set.
We keep $K$ fixed across optimizer LLMs for each dataset. 
% Additionally, following ClusterFusion, we set the clustering sample size to $10 \times K$ in all experiments. 
% \moin{is 10K correct here?yes}
To keep the diagnostic report focused on prominent recurring patterns, we include only clusters whose size exceeds $10\%$ of the total diagnosis pool.

\paragraph{Baseline configuration.}
ACE, GEPA, and MIPRO are run with the same target model and task splits as \method. 
For each baseline, we follow the settings in the original paper or released implementation when available.

For MIPRO, we use the official DSPy\footnote{https://dspy.ai/api/optimizers/MIPROv2/} implementation of MIPROv2, which jointly optimizes instructions and few-shot demonstrations, with \texttt{auto="heavy"} to maximize optimization performance. 
For GEPA, we use the released implementation and adapt the AIME configuration, scaling the optimization budget according to the relative train/development set sizes in our tasks. 
For ACE, we use the released repository for Formula and adapt its instructions to the other datasets. We keep the default setting of one epoch for all tasks\footnote{https://github.com/ace-agent/ace}. 
% \moin{cite or footnote link of repo}

For GEPA, we additionally evaluate a confidence-aware variant in which confidence and calibration diagnostics are provided as auxiliary side information to the reflection prompt, while prompt selection remains driven primarily by task performance.

\subsection{Failure-Mode Persistence}
\label{app:failure_persistence}

We measure the persistence of each failure topic by its average run length, defined as the number of consecutive iterations in which the topic remains active. Longer runs indicate failures that persist despite prompt revisions, while shorter runs suggest failures that are more transient or easier to address.
% Figure~\ref{fig:persistence_all} shows that persistent failures are task-specific. On HotPotQA, the longest-lived issues involve span extraction, surface-form errors, granularity or answer-type mismatches, and multi-hop reasoning. On LiveBench-Math, persistent failures center on arithmetic and algebraic computation, semantic misalignment, and misuse of mathematical definitions or conventions. On Formula, they involve arithmetic calculation, metric definition and formula selection, timing conventions, and domain constraints. Overall, the most persistent failures are not generic formatting errors, but deeper task-specific reasoning failures, suggesting that they often require repeated refinement rather than a single local prompt edit. \moin{or failure mode that the model might not be capable of overcoming simply by prompting (maybe a good segue to the next section if related)}

Figure~\ref{fig:persistence_all} shows that persistent failures are task-specific. On HotPotQA, the longest-lived issues involve span extraction, surface-form errors, granularity or answer-type mismatches, and multi-hop reasoning. On LiveBench-Math, persistent failures center on arithmetic and algebraic computation, semantic misalignment, and misuse of mathematical definitions or conventions. On Formula, they involve arithmetic calculation, metric definition and formula selection, timing conventions, and domain constraints. Overall, the most persistent failures are not generic formatting errors, but deeper task-specific reasoning failures. This suggests that some failures require repeated refinement, and others may reflect limitations that are difficult to overcome through prompting alone.

\subsection{Actionability of Diagnosed Failure Modes}
\label{app:failure_actionability}

\begin{figure*}[t]
    \centering
    \includegraphics[width=\textwidth]{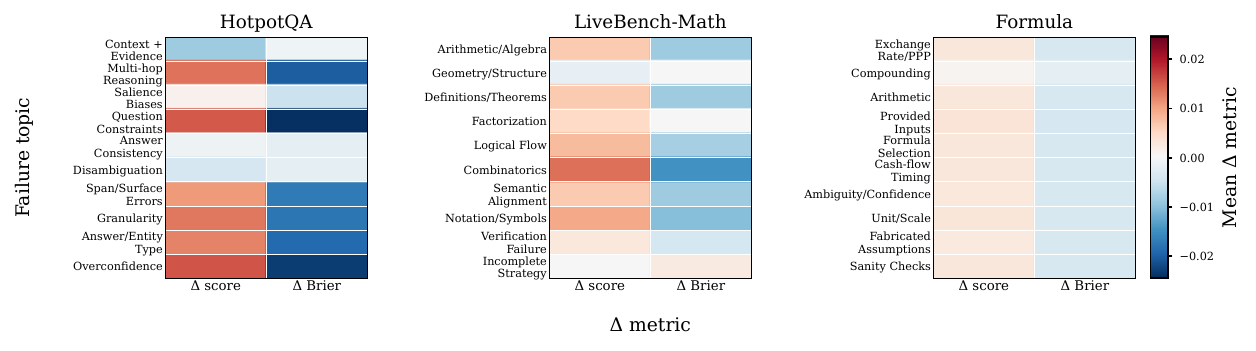}
    \caption{Average next-iteration metric changes associated with each diagnosed failure topic for HotPotQA, LiveBench-Math, and Formula. For each failure topic present at iteration $t$, we report the mean change in task score and Brier score from $t$ to $t+1$. Higher $\Delta$ task score and lower $\Delta$ Brier indicate improvement.}
    \label{fig:failure_delta_all}
\end{figure*}

We analyze which diagnosed failure modes are most actionable for the optimizer. For each failure topic active at iteration $t$, we compute the average change in task score and Brier score after evaluating the revised prompt at iteration $t+1$. This analysis is associative rather than causal, but indicates which diagnosed failures tend to precede useful prompt updates.

Figure~\ref{fig:failure_delta_all} shows that actionable diagnoses vary by task. On HotPotQA, failures involving multi-hop reasoning, question-cue interpretation, span extraction, and granularity mismatches are followed by some of the largest task-score gains and Brier reductions, suggesting that concrete answer-selection errors can often be translated into effective prompt edits. On LiveBench-Math, diagnoses such as combinatorial errors, notation confusion, logical-flow failures, and arithmetic or theorem-application errors are generally followed by task-score gains, though persistent mathematical failures often require multiple revisions.

On Formula, failure topics are less discriminative: most are followed by small task-score gains and Brier reductions, likely because domain-specific failures often co-occur.
Overall, \method is most effective when diagnoses can be converted into explicit behavioral constraints, such as answer-span control, verification steps, or unit and format handling.
More complex mathematical or domain-convention failures remain useful to diagnose, but may require complementary interventions beyond prompt revision.

\subsection{Prompt Length and Development Performance}
\label{app:prompt_length}

\begin{figure*}[t]
    \centering
    \includegraphics[width=0.9\textwidth]{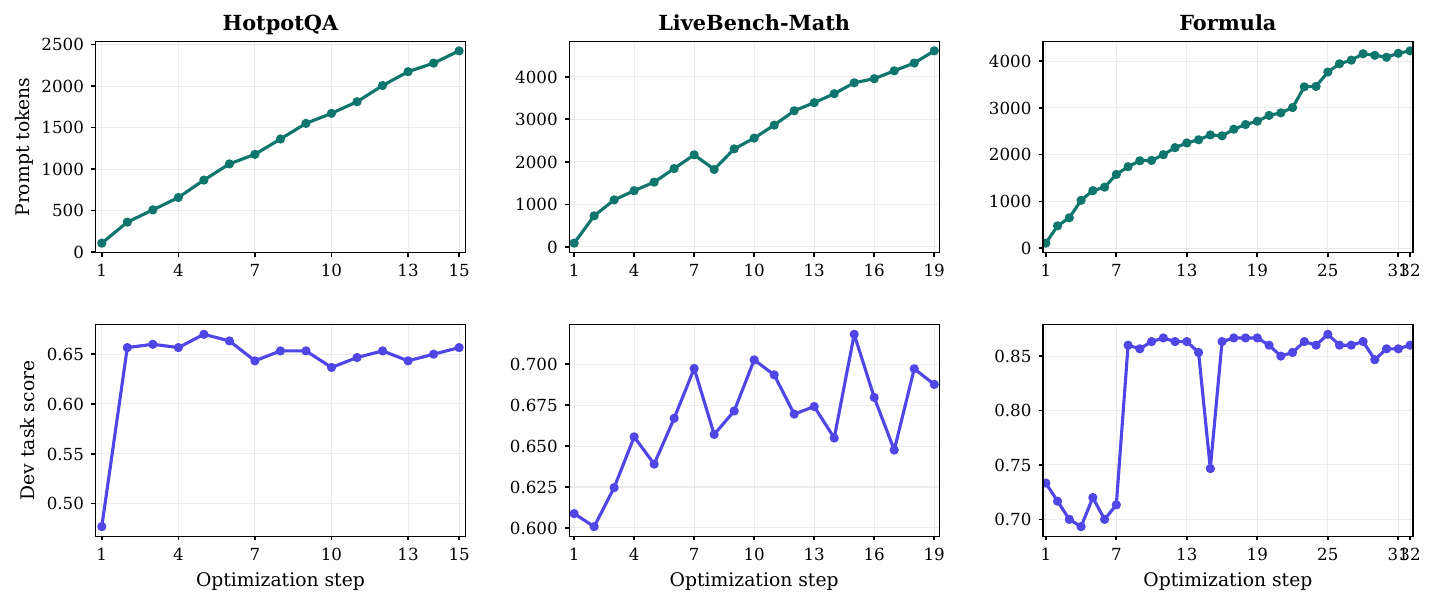}
    \caption{Prompt length and development-set task score across \method optimization iterations for HotPotQA, LiveBench-Math, and Formula. Prompt length generally increases over time, while development performance improves early and then plateaus or fluctuates, motivating development-set prompt selection rather than selecting the final iteration by default.}
    \label{fig:prompt_length_dev}
\end{figure*}

We analyze how prompt length changes during \method optimization and how these changes relate to development-set performance.
Figure~\ref{fig:prompt_length_dev} plots the number of prompt tokens and the corresponding development task score across optimization iterations for HotPotQA, LiveBench-Math, and Formula.

Across all tasks, prompt length tends to increase as the optimizer incorporates additional constraints, checks, and task-specific instructions.
However, development performance does not increase monotonically with prompt length.
On HotPotQA, the largest gain occurs early, after which performance remains relatively stable while prompt length continues to grow.
On LiveBench-Math, dev performance improves overall but fluctuates substantially, with later, longer prompts not always outperforming earlier shorter ones.
On Formula, performance jumps after early revisions and then largely plateaus, despite continued prompt growth.

These trends suggest that longer prompts are not inherently better.
Instead, useful prompt growth comes from adding targeted constraints, while later edits may introduce redundancy or task-specific overfitting.
This further motivates selecting the final prompt by development-set performance, rather than simply using the last prompt produced by the optimization loop.

\subsection{Example Prompt Revision}
\label{app:prompt_revision_example}

Below, we show an example prompt revision produced by \method for HotPotQA with GPT-5 as the optimizer. The initial prompt contains only high-level instructions to reason from context, verify support, and produce a JSON output. The selected prompt is more targeted: it adds explicit guidance for question parsing, supporting-span extraction, answer-type and granularity matching, multi-hop reasoning, temporal or comparative constraints, and confidence calibration.

This example illustrates how \method converts diagnostic feedback into concrete prompt edits. The optimized prompt targets recurring HotPotQA failures by adding explicit controls for minimal span extraction, surface-form matching, multi-hop relation tracing, and confidence calibration under ambiguity.

\begin{example}{Original Prompt - HotpotQA}

\textbf{System Message.}
You are tasked with answering questions using only the provided context.

\textbf{User Instruction.}
\begin{itemize}
    \item Reason step by step using only the provided context.
    \item Be concise but thorough in your justification.
    \item Before answering, verify that your answer is supported by the context.
\end{itemize}

Your output should be a JSON with fields:
\begin{itemize}
    \item justification: a context-grounded explanation of how you reached the answer.
    \item answer: your concise final answer.
    \item confidence: a number in [0,1] representing your confidence in the final answer.
\end{itemize}
\end{example}

\break
\begin{example}{Optimized Prompt - HotpotQA}

\textbf{System Message.}
You are tasked with answering questions using only the provided context.
\textbf{User Instruction.}
Instructions:
\begin{itemize}
    \item Use only the provided context. Parse the question to identify: (a) the subject, (b) the relation/attribute requested, (c) any constraints (time/comparative qualifiers such as based on, before/after, first/last/only), and (d) the required answer type and granularity/format, including whether a unit or administrative/institutional qualifier is explicitly requested.

    \item Find the \hl{specific supporting span(s)} in the context. \hl{When multiple surface forms/aliases exist, select the exact contiguous span in the context that best matches the question wording}; if none match, use the canonical full form as written in context, preserving capitalization, diacritics, and suffixes/post-nominals (e.g., Jr., Sr., III). If the question asks for a unit or an administrative/institutional qualifier, select the span that includes it; if not, and both prefixed and bare forms appear, prefer the bare form.

    \item Answer extraction: \hl{copy the minimal contiguous span verbatim from the context} (casing and numerals). Do not add extra words, qualifiers, punctuation, or units unless explicitly requested. Match the question’s granularity. For numbers, output digits only; if the context has modifiers like ``about/circa/approximately 100'', return just the bare value ``100''. If a full date is given but the question asks for a year, return only the 4-digit year. If the question requests a value-with-unit, copy the exact number+unit as one contiguous span (e.g., ``5 kilometers''). For place/organization names, include prefixes like ``City of'', ``County of'', ``Province of'', ``University of'', or ``Department of'' only if (a) the question targets that unit, or (b) the context consistently uses that prefixed form without an alternative bare form. Trim leading/trailing articles or punctuation unless they are part of the official name.

    \item Yes/No questions: answer exactly yes or no in lowercase with no punctuation.

    \item Relation/compositional queries: \hl{follow all required hops and return the final requested attribute/value}, not an intermediate entity or related item. Verify subject--object direction (e.g., ``X acquired Y'' vs ``X was acquired by Y''). Multi-hop checklist: (1) find the bridge entity tied to the subject, (2) locate the next hop that yields the requested attribute, (3) ensure the final answer span appears verbatim in the context.

    \item Comparative/temporal qualifiers: when prompts include first/last/only/earliest/latest/highest/lowest or a time restriction, (1) collect all candidate spans linked to the subject, (2) apply the specified ordering/constraint to choose the single correct candidate, (3) return only that final span.

    \item Disambiguation: if multiple candidate entities/mentions exist (including same-named entities), choose the one whose qualifiers (role, location, dates, affiliation) satisfy the question’s constraints; ignore distractors.

    \item Before answering, run a \hl{final span checklist}: (a) type and administrative level match the request, (b) temporal/comparative constraint satisfied, (c) surface form matches the question when available, else canonical full form in context, (d) span is minimal, contiguous, and free of extraneous punctuation.

    \item Justification: briefly cite the key supporting phrase(s) and reasoning in $\leq 3$ sentences.

    \item \hl{Confidence calibration}: set confidence based on evidence and ambiguity.
    \begin{itemize}
        \item 0.85 if exactly one unambiguous supporting span directly answers the question with no alias/format ambiguity.
        \item $\sim$0.55 if supported but there are multiple aliases/forms, qualifier/unit risks, or minor ambiguity (cap confidence here in such cases).
        \item $\sim$0.35 if partially supported or requires weaker inferences across snippets.
        \item $\leq$0.25 if the support is tenuous or potentially contradicted.
    \end{itemize}
\end{itemize}

Output JSON fields:
\begin{itemize}
    \item justification: a context-grounded explanation of how you reached the answer.
    \item answer: the bare minimal span only (no quotes or extra words).
    \item confidence: a number in [0,1] representing your confidence in the final answer.
\end{itemize}

\end{example}

\end{document}